\documentclass[conference]{IEEEtran}
\IEEEoverridecommandlockouts
\usepackage{cite}
\usepackage{amsmath,amssymb,amsfonts, amsthm}

\usepackage{caption}
\usepackage{subcaption}

\usepackage{comment}
\usepackage{array}
\usepackage{booktabs}
\newcommand{\etal}{{\em et al.}}
\theoremstyle{definition}

\renewcommand{\thedefinition}

\newcounter{inlineenum}
\renewcommand{\theinlineenum}{\Roman{inlineenum}}

\usepackage{algorithmic}
\usepackage{graphicx}
\usepackage{textcomp}
\usepackage{xcolor}
\def\BibTeX{{\rm B\kern-.05em{\sc i\kern-.025em b}\kern-.08em
    T\kern-.1667em\lower.7ex\hbox{E}\kern-.125emX}}

\begin{document}

\title{How much data do I need?\\ A case study on medical data}

\author{\IEEEauthorblockN{Ayse Betul Cengiz}
\IEEEauthorblockA{\textit{School of Computing} \\
\textit{Newcastle University}\\
Newcastle upon Tyne, UK \\
a.b.cengiz2@newcastle.ac.uk}
\and
\IEEEauthorblockN{A. Stephen McGough}
\IEEEauthorblockA{\textit{School of Computing} \\
\textit{Newcastle University}\\
Newcastle upon Tyne, UK \\
stephen.mcgough@ncl.ac.uk}

}

\maketitle
\IEEEpubidadjcol
\begin{abstract}
The collection of data to train a Deep Learning network is costly in terms of effort and resources. In many cases, especially in a medical context, it may have detrimental impacts. Such as requiring invasive medical procedures or processes which could in themselves cause medical harm. However, Deep Learning is seen as a data hungry method. Here, we look at two commonly held adages i) more data gives better results and ii) transfer learning will aid you when you don't have enough data. These are widely assumed to be true and used as evidence for choosing how to solve a problem when Deep Learning is involved. We evaluate six medical datasets and six general datasets. Training a ResNet18 network on varying subsets of these datasets to evaluate `more data gives better results'. We take eleven of these datasets as the sources for Transfer Learning on subsets of the twelfth dataset -- Chest -- in order to determine whether Transfer Learning is universally beneficial. We go further to see whether multi-stage Transfer Learning provides a consistent benefit. Our analysis shows that the real situation is more complex than these simple adages -- more data could lead to a case of diminishing returns and an incorrect choice of dataset for transfer learning can lead to worse performance, with datasets which we would consider highly similar to the Chest dataset giving worse results than datasets which are more dissimilar. Multi-stage transfer learning likewise reveals complex relationships between datasets.

\end{abstract}


\section{Introduction}
The modern world is data focused. Almost any task which is performed these days generates data. This is no more true than in the area of medicine. From the humble digital X-Ray through to MRI scans to the processing biopsies for rare medical conditions~\cite{atif}. From this wealth of data we are able to train Machine Learning (ML) approaches to aid us in the detection, diagnosis and treatment of many conditions. The naive assumption here is that the more data the better. However, this is not always best, or possible. It may be that the cost of such a procedure makes such data collection less than desirable. More significantly, in the case of medicine, the data collection is not desirable -- to produce a data sample for lung cancer, for example, requires someone to have lung cancer. Therefore we would prefer not to have the person to have lung cancer. Also, a number of medical data collection processes are quite invasive -- requiring for example biopsies -- which reduces the number of people willing to participate in such data collection. Or the medical process to collect the data could have its own risk for the patient.

If we focus here on Deep Learning solutions the volume of data required is far larger -- leading to situations where developments can only happen for common conditions where data collection is cheap, easy and safe. However, there is a strong drive to move beyond these `easy' conditions and democratise the benefits of Deep Learning to all conditions.

There are three adages which are often heard within the ML community:
\begin{itemize}
    \item More data gives better results.
    \item Transfer learning will aid you when you don't have enough data.
    \item Data augmentation will aid you when you don't have enough data.
\end{itemize}

All of these would appear to hold great hope for the medical community where data is often scarce. In this work we unpick the first two of these, leaving the third as future work. We analyse if these adages are true, and if so can we make a more definitive rule or rule-set. Our focus in this work is on Deep Learning (DL) models as these are more data hungry than other approaches.

The adage that more data will lead to better results is a compelling one. It would seem intuitive that more data is better. However, with DL the relationship between dataset size and model performance is not simple. The `quality' and `variety' of the data are often more important. We therefore evaluate, for a number of datasets, how the performance changes as the dataset volume increases.

Transfer Learning (TL)~\cite{b2} where a model trained on a source domain before fine-tuning on a (small) volume of the target domain has been demonstrated to be highly applicable in many DL tasks. TL is a Deep Learning technique for applications that lack sufficient training data. The model's ability, learned from one task, is transferred in the form of model weights which are used as a starting point when fine-tuning for a different domain. Intuitively, it is expected that the best results might be obtained from a similar source domain to the target domain. Here we loosely define similar, as it is an open research question as to how similarity is defined for Deep Learning. For example, it would be assumed that a dataset for identifying stray cats from domesticated cats could be used as the source domain for a task in identifying different breeds of pedigree cats. As many of the core features extracted by the Deep Learning network for both cases could be assumed to be similar.

It has been proposed that multiple transfer learning steps -- known as Multi-stage Transfer Learning (MSTL)~\cite{b3, b4} may yield better results. The idea being to fine-tune the model multiple times from less similar to more similar domains normally on progressively smaller datasets.

Here, a crucial question arises: what does similarity between domains mean? It is worth noting that the human perspective and the machine perspective are quite different. Generally, humans define similar objects according to their common features such as parts or limbs, colours, and shapes. Humans look at the images in the dataset by what they are~\cite{b5}. However, Neural networks see images without knowing what they are, instead, they consider several low-level features such as edges, curves, lighting, and colour scheme, among images~\cite{b6}.

From our work we identify that for more data this holds strongly until the dataset is over a certain size, beyond this point improvements are possible but it becomes a case of diminishing returns. Transfer Learning is also a much more complex scenario where it is possible to make good and bad choices as to which datasets we transfer from. In many cases transfer learning can lead to worse results than would be obtained if we didn't use it. Likewise, when doing multi-stage transfer learning the choice of the set of datasets to transfer from is a more complex proposition. We look at transferring from very general datasets through to highly similar (to the target domain) datasets and find that this is not necessarily the best approach to take. 

The rest of the paper is organised as follows. In Section \ref{sec:lr} we discuss the prior literature in the areas of data volume and Transfer Learning. Section \ref{sec:method} presents our experiments before we present the results of these experiments in Section \ref{sec:results}. We bring these threads together in our conclusions -- Section \ref{sec:conc} -- along with proposing directions for future work.


\begin{table*}[tb]
  \centering
  \caption{The datasets used}
  \label{tab:datasets}
  \begin{tabular}{>{\raggedright}p{2cm}c>{\raggedright}p{12cm}c@{}}
    \hline
    \textbf{Name} & \textbf{Type} & \textbf{Description} & \textbf{Classes} \\
    \hline
    ImageNet\cite{b12} & Natural & A total of 1,000,000 images, 1,000 from each category. & 1000 \\
    Imagenette\cite{b20} & Natural & 10 easily classified classes from ImageNet (tench, English springer, cassette player, chain saw, church, French horn, garbage truck, gas pump, golf ball \& parachute). 1350 images per class & 10 \\
    Natural\cite{b21} & Natural & 6,899 image samples of airplanes (727), cars (968), cats (885), dogs (702), flowers (843), fruits (1000), motorbikes (788) \& people (986). & 8 \\
    Imagewoof\cite{b20} & Natural & About 28,000 medium-quality animal images. Images per class varies from 2,000 to 5,000. & 10 \\
    Animal\cite{b22} & Natural & 10 animal classes (dog, cat, horse, spider, butterfly, chicken, sheep, cow, squirrel, elephant). Images per class varies from 2,000 to 5,000. & 10 \\
    Pet\cite{b23} & Natural & 25,000 images of cats (12,491), and dogs (12,470). & 2 \\
    Chest\cite{b24} & Medical & A total of 5,863 images. Chest X-rays in patients with pneumonia. & 2 \\
    Covid\cite{b25,b26} & Medical & Chest X-ray images (COVID-19 (3,616), Lung Opacity (6,012), Viral Pneumonia (1,345), and Normal (10,200)) & 4 \\
    Tuberculosis\cite{b27} & Medical & Chest Xray images (Normal (3,500), and Tuberculosis (700)) & 2 \\
    Kidney\cite{b28} & Medical & Kidney CT images (Cyst (3,709), Normal (5,077), Stone(1,277), \& Tumor (2,283)) & 4 \\
    Breast\cite{b29} & Medical & Breast Ultrasound Scans (bening (891), malignant(421) \& normal(266)) & 3 \\
    Mura\cite{b30} & Medical & Upper limb bone Xrays (normal (23,602) \& abnormal (16,412)) & 3 \\
    \hline
  \end{tabular}
\end{table*}

\section{Literature review}
\label{sec:lr}
Our Literature Review navigates through two crucial aspects of Deep Learning: the influence of Data Volume and the power of Transfer Learning. We begin by exploring how the amount of data affects performance in Deep Learning networks. Then, we dive into Transfer Learning, examining its relevance in fields with limited data, such as medical image classification.
\subsection{Data volume}

Given the significance of data volume on the performance achieved from a Deep Learning network there have been, to date, relatively few studies analysing the issue. 

Cho \etal~\cite{cho} illustrate accuracy versus data volume curves for CT scans, focusing on different body parts. They also identify the steep initial improvement as data volume increases and then a diminishing return after a certain volume. We go further in this work and demonstrate this effect on twelve different datasets.

Work by Sun \etal \cite{sun} shows only a linear relationship between data volume and performance. However, we would argue as their smallest data volume is ten million samples they have started their analysis from too large a volume.

\subsection{Transfer Learning}

Here, we present studies examining the transfer of information between different deep networks, especially in data-scarce fields.

Transfer Learning (TL) refers to a scenario where the features learned for one task are leveraged to improve the generalization in another task. Considering images as an example, there are several shared low-level features, such as edges, curves, lighting, and colour patters, common among images \cite{b6}. Hence, the weights of a model trained on one dataset can be applied to another. To exploit these benefits, especially in the field of medical image classification (MIC), we can utilise pre-trained CNNs using the ImageNet \cite{b12} dataset (see \cite{b10, b11}). 

TL can be applied more than once, with the intermediary datasets often referred to as a bridge datasets. Kim \etal \cite{b3} applied multi-stage TL to X-ray, MRI, and CT images. They used ImageNet as the source dataset for these three types of images. Claiming an increase in the performance of the DL model by using a bridge dataset similar to the target dataset. However, in this study, the researchers only use ImageNet as a source. Therefore, the source and target (since it is a medical dataset) are always irrelevant areas. We focus on every possible combination.

Orenstain \etal \cite{b13} evaluated the effects of out-of-domain data on classification accuracy using three datasets containing plankton images collected with different tools and techniques. As a result of their evaluation, training the output layer twice with different datasets had a positive effect on the classification of the target data set. The main drawback here is that all three datasets they use contain different schemata of the same objects. Therefore, it is not as comprehensive a study as we aimed for.

Alkhaleefah \etal~\cite{b14} proposed TL, which updates the learnable parameters of any pre-trained network by modifying their dense layers with a large dataset that is similar to the target dataset, using other publicly available datasets in the same domain in order to classify benign and malignant breast cancer with a very limited dataset. Zhan \etal~\cite{b15} calibrated a pre-trained ResNet34 model with ImageNet with the pneumonia classification chest x-ray dataset as a bridge dataset and finally used the calibrated model for the detection of COVID-19 disease in the target dataset. Godasu \etal~\cite{b16} proposed a multi-stage learning system by using the pneumonia classification dataset Chestxray14 \cite{b17} as the bridge dataset, to detect hemorrhages from head CT images. Matos \etal~\cite{b18} used two source datasets for covid-19 disease detection from x-ray images, one of which was a different domain and the other was the same domain as the target dataset. The outcomes of the suggested method have demonstrated that it is feasible to use information from other datasets by transferring it more than one time to increase the accuracy of a specific classification task. Asigwa \etal \cite{b19} introduced multi-level transfer learning, which involves combining intermediate stages of transfer learning from related auxiliary domains to combat performance loss when the target domain is constrained, and the domain and target are unrelated. Likewise, although these studies are in the field of medicine, the source dataset and intermediate datasets are very close datasets to the target dataset.

\section{Methodology}
\label{sec:method}
We investigate various datasets and how subsets of these datasets affect the performance under Deep Learning and Transfer Learning. We first discuss the datasets and then go on to investigate how the accuracy of a ResNet18~\cite{resnet18} is affected by taking different proportions of the dataset. We then cover the search space of Transfer Learning which we evaluate for different source and intermediate datasets.

Given that the practitioner does not have access to an unlimited data supply we conduct three experiments below to ascertain the best policy for them to adopt. Based on either collecting more data or performing single / multi-stage TL.

\subsubsection{Data volume}
We add to the conversation as to how much data is needed. For each dataset we randomly sample $p\%$ of the dataset and perform a full training of a ResNet18 network. We conduct each sample-training combination 20 times to reduce bias from dataset sampling. Although other networks may yeild better results we stick with ResNet18 for consistency expecting the same pattern to occur with other networks. 

\subsubsection{Single stage TL}
We instead look at how a network pre-trained on one of our datasets (referred to as the source) and then fine-tuned using TL on a proportion of the target dataset could benefit our underlying challenge. We investigate how similarity of the datasets affects the performance of the challenge.

\subsubsection{Multi-stage TL}
Here we expand the previous experiment to have multiple TL stages and evaluate how the similarity of the data in each stage affects the overall performance. Is it better to move from general datasets through similar datasets and then finally the most similar dataset or vice versa.

\begin{figure*}[]
     \centering
     \begin{subfigure}[b]{\textwidth}
         \centering
         \includegraphics[width=\textwidth, trim={2.5cm 8.5cm 3.5cm 8cm}, clip]{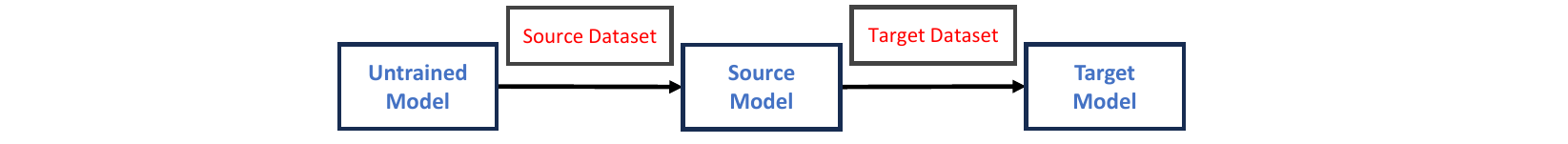}
         \caption{Simple Transfer Learning}
         \label{subfig:simpleTL}
     \end{subfigure}
     \hfill
     \begin{subfigure}[b]{\textwidth}
         \centering
         \includegraphics[width=\textwidth, trim={3cm 8.5cm 4cm 8cm}, clip]{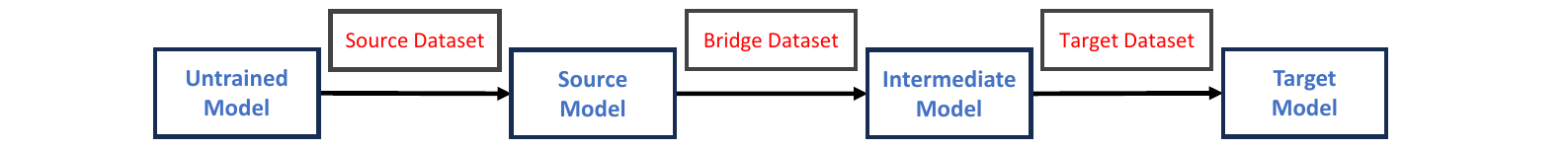}
         \caption{2-bridge MSTL}
         \label{subfig:1bmstl}
     \end{subfigure}
     \hfill
     \begin{subfigure}[b]{\textwidth}
         \centering
         \includegraphics[width=\textwidth, trim={2.8cm 8cm 3.5cm 8cm}, clip]{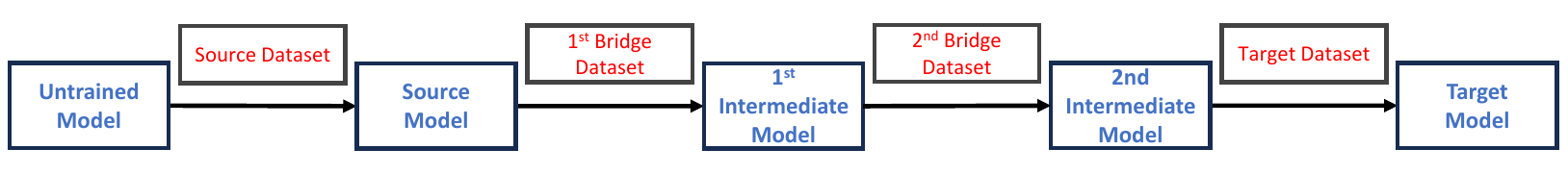}
         \caption{3-bridge MSTL}
         \label{subfig:2bmstl}
     \end{subfigure}
        \caption{Stages of MSTL}
        \label{fig:MSTL}
\end{figure*}
\subsection{Datasets}
\label{subsec:datasets}
We use twelve datasets in this work (Table \ref{tab:datasets}) comprising of six general datasets (\textit{Natural}), which are considered to be very different from the target dataset (Chest), and six medical datasets, which have different levels of similarity to the target dataset. We hold the same 20\% of the data back for all testing. We create sample sets from the remaining 80\% comprising of p\% of the dataset to reflect smaller datasets using the same sample set for each experiment to ensure fair comparison. 

Formally if $D=\{x_1, ..., x_N\}$ is the collection of $N$ elements in the dataset, then we pick $pN (0 \leq p \leq 1)$ indices at random from $N$ ($r_1, ..., r_{pN}$) such that $0 \leq r_i \leq N$ and $\nexists i,j$ s.t. $r_i = r_j$ and $i \neq j$. 

\subsection{DL Architecture}
We use ResNet18 as our sample model as compared to other CNN architectures ResNet18 produces a high accuracy with low computational cost due to having fewer parameters \cite{b33} -- for example reaching over 98\% accuracy on the Breast Cancer dataset{\color{red}} \cite{b35}. Although other architectures could achieve higher accuracy we are more interested here in the relative accuracy when varying the amount of training data. We would expect the findings of this work to be replicate for other architectures, but see this as part of our future work. We use the default implementation of ResNet18 from PyTorch with no parameter changes. Using the pretrained ImageNet v1 weights from PyTorch when used. For consistency each model was trained for 10 epochs.

\subsection{Data Volume}
\label{subsec:data_volume}
We vary here the percentage of data used for training a ResNet18 network on each of our datasets between 5 and 100\% in steps of 5. In each case we re-run the experiment twenty times. From this we hope to ascertain a relationship between the size of training dataset and the performance achieved. If the adage is true we would expect to see an increasing performance as the dataset size increases.

\begin{figure*}[ht]    \centerline{\includegraphics[width=\linewidth]{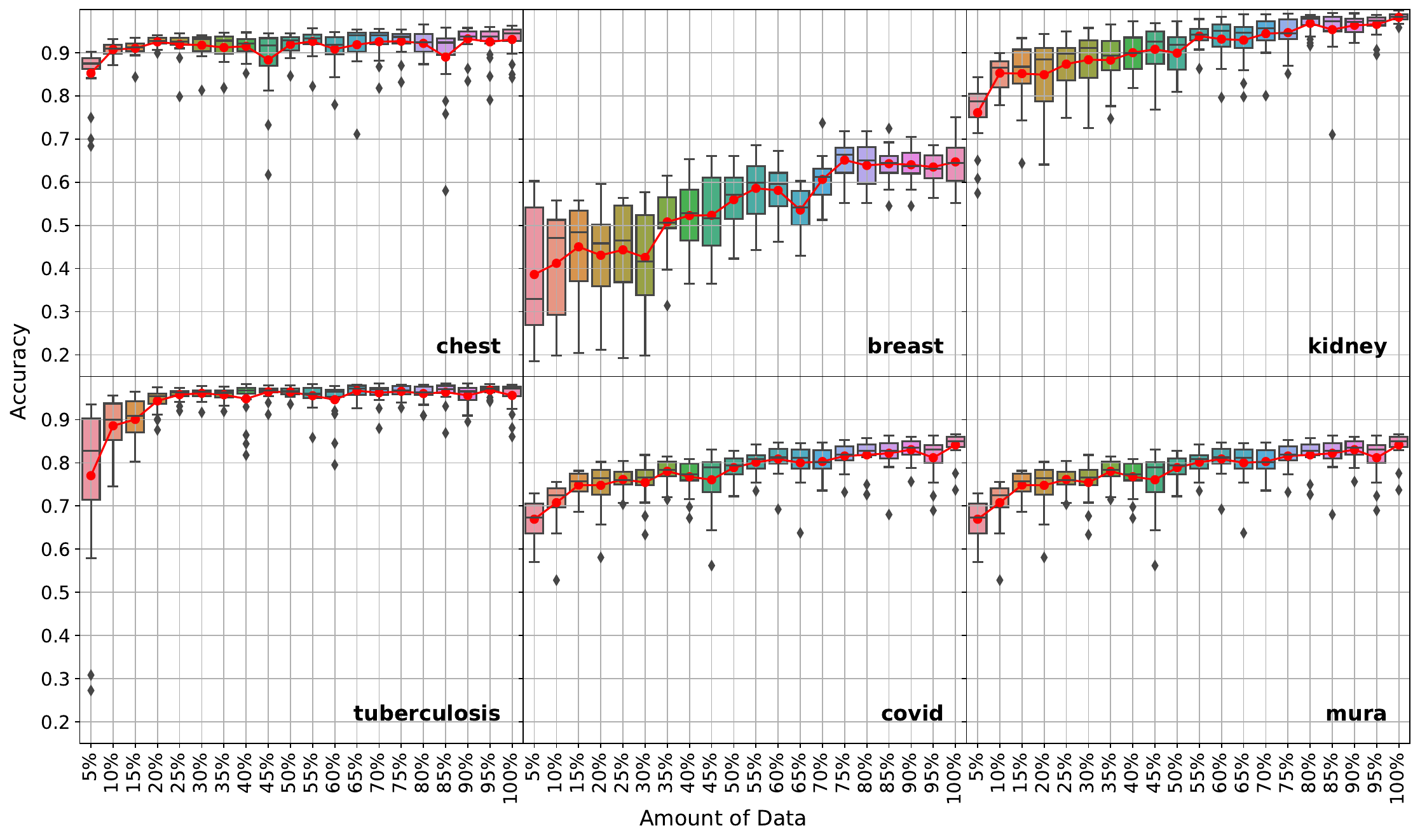}}
    \caption{Inference Results from training different percentages of data on ResNet18}
    \label{fig:basic18}
\end{figure*}

\subsection{Single stage TL}
\label{subsec:single_TL}
In this case (Figure~\ref{subfig:simpleTL}), the \textit{Untrained Model} is trained by the source dataset to create the \textit{Source Model}. After training the \textit{Source Model} further training is performed using the target dataset to create the \textit{Target Model}. This is the standard approach for TL~\cite{b31}. In our experiments Chest is the target dataset and the other eleven datasets are our different source datasets. Here we focused on two adaptation techniques: feature extraction and fine-tuning \cite{b32, b33}. When training with the target dataset we can choose which layers of the network can be updated and which will remain fixed -- often referred to as frozen. We define three different un-/frozen options:

\begin{description}
\item[PG0] all layers are unfrozen (PG is Parameter Group).
\item[PG1] the last convolutional layer and fully connected layers are unfrozen, all others are frozen.
\item[PG2] Only the fully connected layers are unfrozen.
\end{description}

\subsection{Multi stage TL}
\label{subsec:multi_TL}
It has been proposed that better performance can be obtained by performing multiple iterations of transfer learning~\cite{b18}. In this case the single-stage TL is extended with further intermediate datasets (leading to intermediate models). Figures \ref{fig:MSTL}(b) and (c) illustrate this for the cases of two and three stages -- often called 2-bridge and 3-bridge respectively. We wish to evaluate here the naive assumption that one should move from a source dataset which is very general to progressively closer to the target dataset.

With our final target being the Chest dataset we perform two experiments:

A 2-bridge TL approach where the first dataset is either \textit{Natural} (excluding ImageNet) or \textit{Medical} (exclusing Chest) and the second bridge dataset is the opposite. Thus 25 experiments.

A 3-bridge TL approach where the first dataset is ImageNet with the second and third datasets being \textit{Natural} and \textit{Medical} in the same alternatives as 2-bridge. Again 25 experiments.

In all cases the final (target) training was on Chest with the other eleven datasets forming the prior training sets -- natural datasets forming the most general datasets and the medical datasets being similar to Chest. 

\subsection{Varying the Amount of Data for TL}
\label{subsec:data_amount}
We also evaluate how the volume of data available affects the TL process. This is done with two types of experiment:

{\bf Only vary target dataset} with all other datasets being used in full. Here p\% varied between 5\% and 100\% in steps of 5 for the single-stage TL. However for the multi-stage TL we only  evaluate up to 20\% as this was demonstrated to be the most interesting region for Chest in terms of change in performance.

{\bf Vary all datasets} with each dataset (source, intermediate and target) being reduced to a fixed number of images -- $nC$ -- where $n \in [120, 240, 480, 960]$ and $C$ is the number of classes in that dataset. 

\section{Results}
\label{sec:results}

All experiments were run on a Nvidia Titan RTX with an Intel(R) Core i9-9820X CPU @ 3.30GHz, and 64GB RAM,  following our methodology.

\subsection{Data volume}
\label{sec:dvExp}
We focus here on the shapes of the curves -- see Figure \ref{fig:basic18} -- as opposed to the specific accuracy's achieved. It should be noted that it is possible to achieve better accuracy with a different Neural Network, however, the argument here is that the shape of the curves would remain largely the same. Interestingly all datasets apart from breast show a hocky-stick shape where below about 20\% the accuracy improves rapidly but after $\sim$20\% the accuracy only improves at a significantly reduced rate. This would seem to suggest that once you have reached an initial minimum viable dataset size then there is a law of diminishing returns on collecting more data. The breast dataset may buck this trend due to its small size in comparison to the other datasets. With the the hockey elbow still to come.

\subsection{Single stage TL}

\begin{figure*}[t]
    \centering
    \resizebox{1.0\linewidth}{!}{\includegraphics{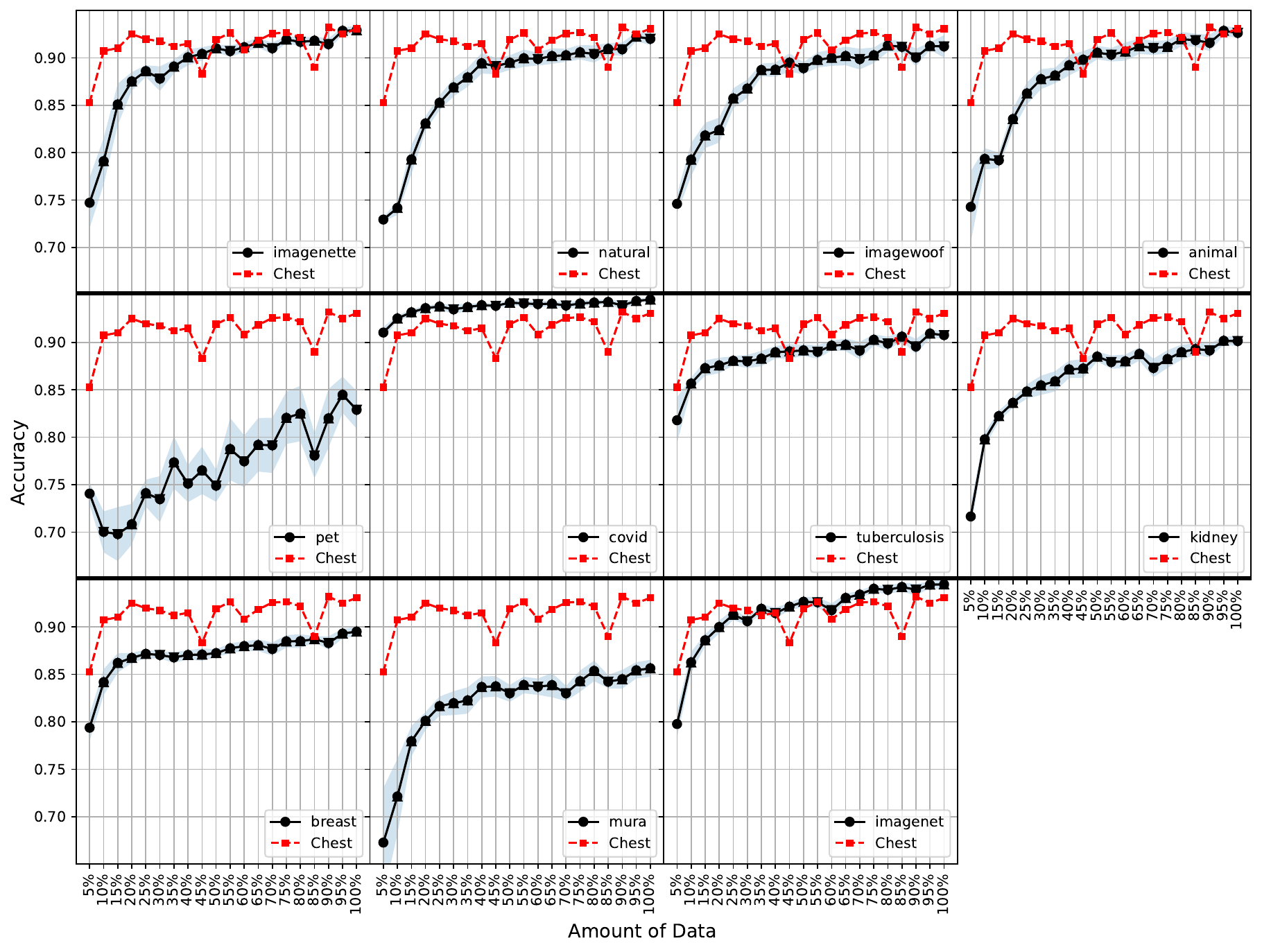}}
    \caption{Inference Results of Chest Domain - Knowledge is Transferred from the Other Datasets}
    \label{fig:TL18}
\end{figure*}
 
{\bf Percentage of Chest:} We train a ResNet18 first on the entirety of one of the datasets (except for Chest) and then TL to the Chest dataset -- varying the percentage of the chest dataset we use. In this case the TL approach used is PG2. Figure \ref{fig:TL18} shows this accuracy (blue), along with 95\% confidence interval, for each dataset when taking p\% of the chest dataset. The results for Chest without TL are also shown (red) for comparison.

Interestingly, the majority of these datasets for TL to Chest do not produce better results in comparison to training the Chest dataset on its own. Covid is the only dataset to always provide superior performance to Chest on its own. This is likely to be due to the close similarity between the two datasets -- both chest X-Ray images. This may lead to the assumption that the more `similar’ a dataset is to the target domain the better. However, tuberculosis -- another dataset of chest X-Ray images -- fails to achieve better results than Chest on its own. Therefore, even if the datasets may appear similar in content type, they may not be beneficial for TL. The results for ImageNet may appear counter-intuitive. When we only have a small volume of data for Chest training directly gives better results than using TL from ImageNet whilst for larger amounts  (approximately over 40\%) of Chest data using ImageNet TL is better than direct training. This may be due to the network being too focused on the ImageNet dataset and a minimum of data from Chest being required to overcome this effect. 

Another interesting case is mura, X-ray images of bones, which is the worst accuracy at 5\% and is eventually beaten by all but pet. Never beating Chest on its own. This would seem to suggest that an extremely similar dataset is best to transfer learn from, though once similarity reduces -- even if only sligtly (still X-Ray) -- TL can become significantly worse.

{\bf Vary all datasets:}

Here the number of images for each class, in both the source and the target dataset was set to $nC$ and we used PG0 for TL. We excluded the brest and ImageNet datasets from this experiment due to too little and too much data respectively. 

\begin{figure*}[t]
    \centerline{\includegraphics[width=\linewidth, trim={0.7cm 0cm 3cm 0cm}, clip]{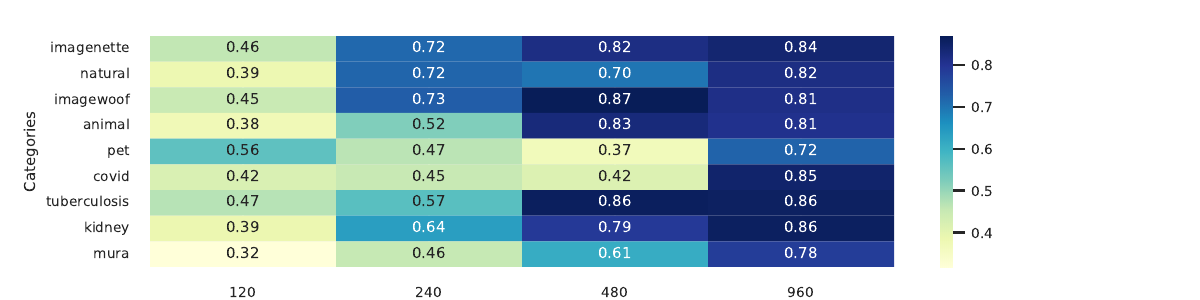}}
    \caption{Accuracy with fixed dataset size (both source and target) for single-stage TL: x-axis values are the volume of data used, y-axis is the source dataset and cell values are the accuracy achieved}
    \label{fig:npg0}
\end{figure*}

\begin{figure*}[t]
    \centerline{\includegraphics[width=0.95\linewidth, trim={0 0 0 0}, clip]{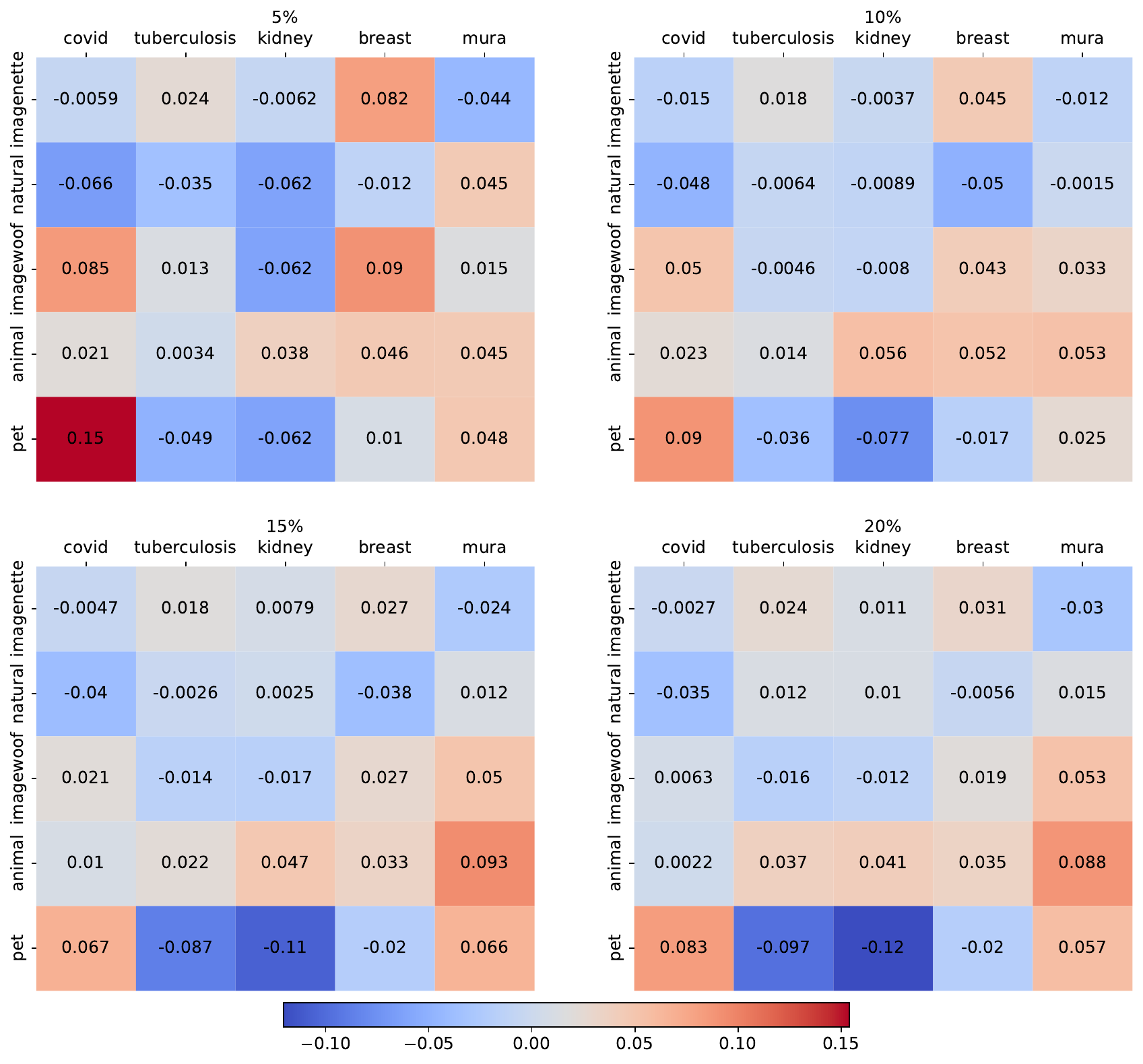}}
    \caption{Comperative Inference Results of 2-Bridge MSTL}
    \label{fig:1_brd_mstl}
\end{figure*}

In general there is a clear relationship between training dataset size and accuracy (Figure \ref{fig:npg0}). Pet bucks this trend, though this could be due to the fact that it is also a two-class problem and thus the fully-connected layers need less fine-tuning. In general the natural datasets do better at lower volumes of data -- potentially due to the fact that they have more classes and hence provide more training data. Covid does particularly badly considering how well it does when the full Covid dataset is used. Conversely, tuberculosis does much better than when the full datset is used. Thus suggesting a more complex relationship between similarity and data volume. And that larger source datasets are beneficial.

\subsection{Multi stage TL}

In these experiments we limit ourselves to two- and three-bridge TL. Arguing that if there is no clear benefit from this number of bridges then moving higher is unlikely to provide any advantage.


\subsubsection{Two-Bridge Transfer Learning}

\ 

{\bf Percentage of Chest:}
As most datasets in Section \ref{sec:dvExp} had started to plateau by about 20\% we limit our experiments to [5\%, 10\%, 150\%, 20\%] of Chest.

The untrained model is initially trained (using PG0) with the entire source dataset. Subsequently, TL is used (with PG2) using the complete bridge dataset. Finally, TL (with PG2) is used with a percentage of the target dataset.

It should be stated that three-bridge TL can lead to better results -- the best three-bridge results yielded an accuracy of $\sim$0.9304 (followed by $\sim$0.925 and $\sim$0.9123, using 20\%, 15\%, and 10\% respectively), whilst the best simple transfer learning yielded $\sim$0.925 (using 20\%).

In Figure \ref{fig:1_brd_mstl} we investigate if a general dataset TL to a similar dataset (c.f. the target dataset) before TL to Chest -- we denote this as {\em general $\rightarrow$ similar $\rightarrow$ Chest} -- yeilds better results than  {\em similar $\rightarrow$ general $\rightarrow$ Chest} in two-Bridge TL. For each cell we compute $a(natural \rightarrow medical \rightarrow Chest) - a(medical \rightarrow natural \rightarrow Chest)$, where $a(\square \rightarrow \triangle \rightarrow Chest)$ is the accuracy from the 2-bridge TL $\square$ followed by $\triangle$ followd by Chest. Thus, a negative (blue) value indicates that {\em similar(medical) $\rightarrow$ general(natural) $\rightarrow$ Chest} yields better accuracy than {\em general $\rightarrow$ similar $\rightarrow$ Chest}.

There is no clear winner on general followed by similar or vice versa. There is a slight advantage of using general followed by similar [16, 15, 16] cases out of 25 for the [5\%, 15\%, 20\%] datasets respectively. In general the difference between the two orderings of datasets becomes less as the proportion of Chest increases. The only case which bucks this trend is {\em kidney $\rightarrow$ pet $\rightarrow$ Chest} which becomes more pronounced as the volume of Chest increases. Though as Kidney is potentially the least similar medical dataset to Chest and may be more equivalent to {\em almost general $\rightarrow$ general $\rightarrow$ Chest} this may explain this situation.

\begin{figure*}[t]
    \centerline{\includegraphics[width=0.95\linewidth, trim={3.5cm 0 4cm 1cm}, clip]{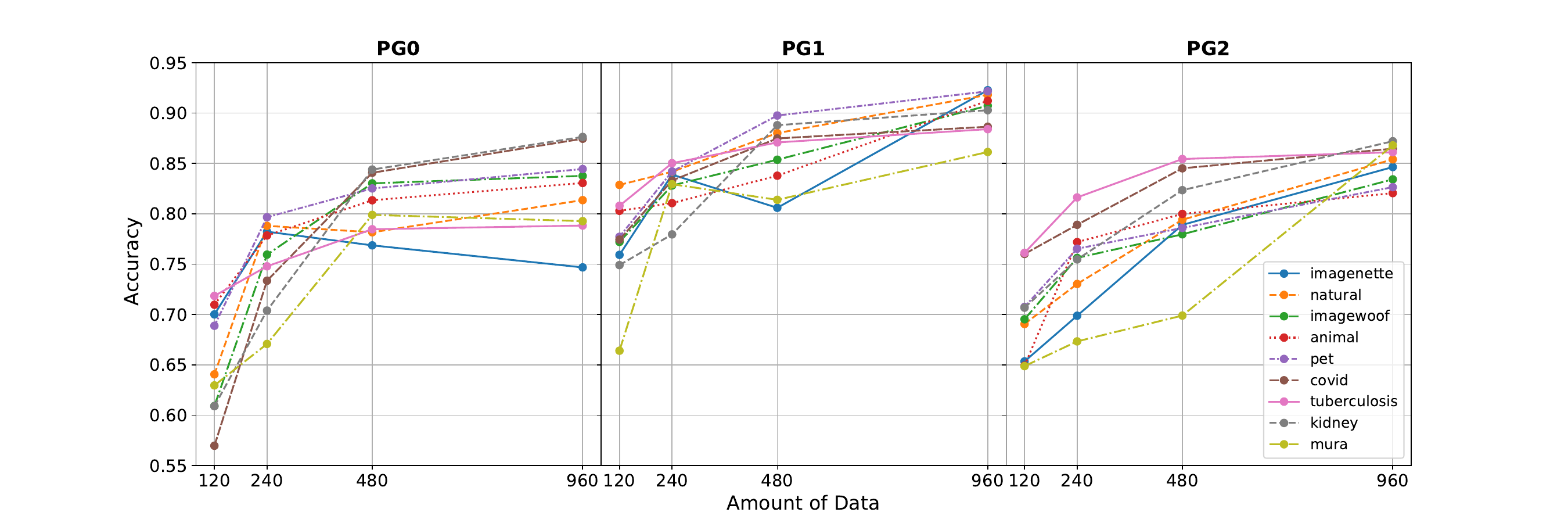}}
    \caption{Accuracy for Constant Data (2-Bridge): x-axis values are the volume of data used, y-axis is the accuracy achieved}
    \label{fig:pg_mstl}
\end{figure*}

{\bf Vary all datasets:}
We used a pre-trained (ImageNet) ResNet18 with a fixed amount of data for both the intermediate and target dataset. I.e., {\em ImageNet $\rightarrow$ intermediate[nC] $\rightarrow$ target[nC]}, where $\square[nC]$ is where we take $nC$ samples for dataset $\square$. We evaluate the difference between PG0, PG1 and PG2 TL approaches in Figure \ref{fig:pg_mstl}.

In general PG1 yielded the best results. Having in general better results at smaller and larger volumes of data. Out of PG0 and PG2, PG2 gives better results -- higher accuracy at low data volumes and more consistent value at higher volumes, though it is possible to obtain higher accuracy with some intermediate datasets with PG0 and larger volumes of data. Thus there is no simple answer which always holds, though in general PG1 is more likely to give you better results.

\subsubsection{Three-Bridge Transfer Learning}

\ 

Using a pre-trained (ImageNet) ResNet18 network we then TL twice with either a general, similar pair of datasets or vice versa with only up to 20\% of the final dataset. Thus {\em ImageNet $\rightarrow \square \rightarrow \triangle \rightarrow$ Chest} where $\square, \triangle$ are either general or similar datasets. The best performing combination was {\em ImageNet $\rightarrow$ ImageNette $\rightarrow$ Covid $\rightarrow$ Chest} which yielded $\sim$93.98\% accuracy at 20\% of Chest data.

\begin{figure*}[ht]
    \centerline{\includegraphics[width=0.95\linewidth, trim={0 0 0 0}, clip]
    {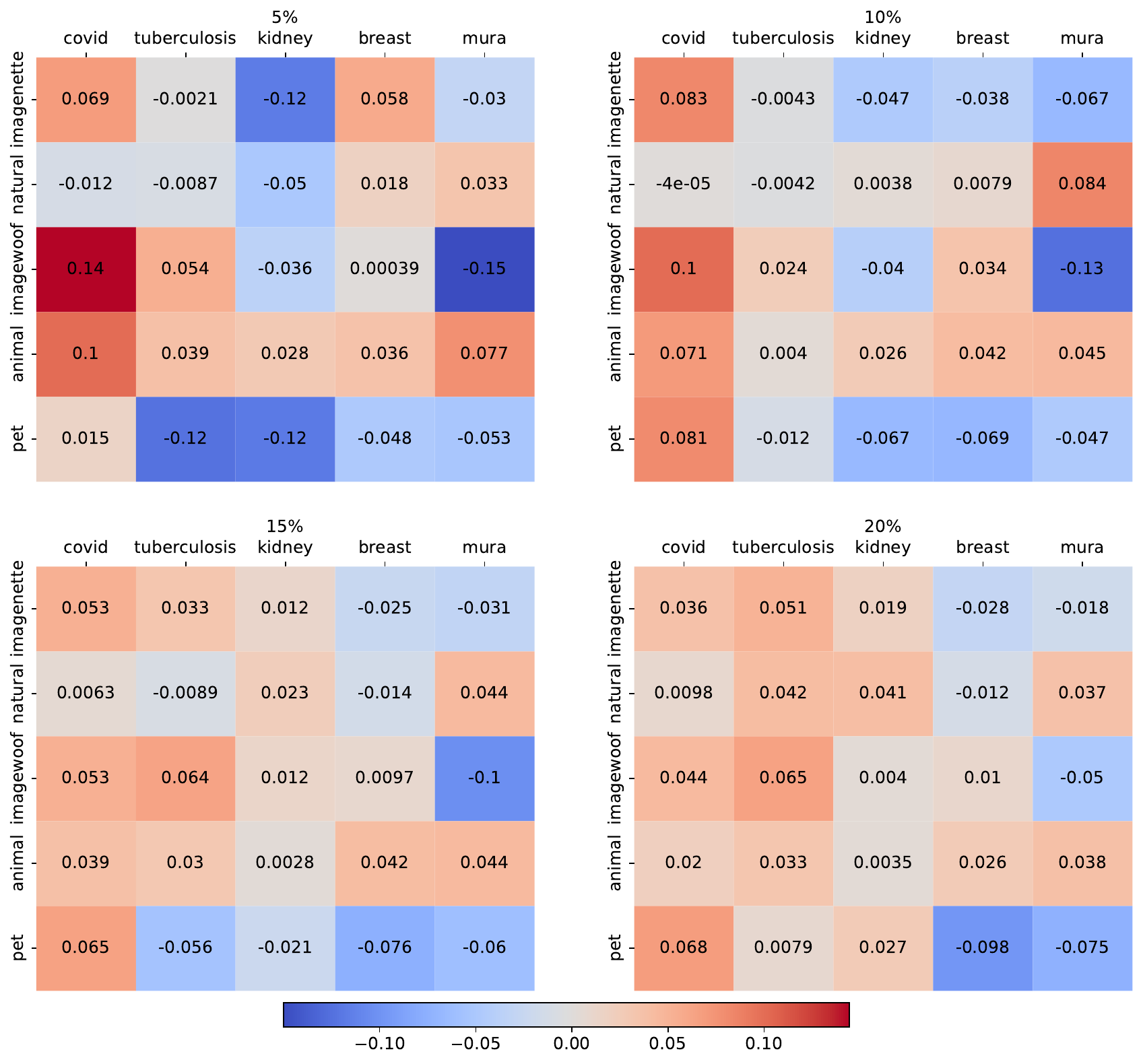}}
    \caption{Comparative Inference Results of 3-Bridge MSTL}
    \label{fig:2_brd_mstl}
    \vskip 8pt
\end{figure*}

We adopt the same figure construction approach as Figure \ref{fig:1_brd_mstl} for the first and second intermediary datasets. I.e., $a(ImageNet \rightarrow natural \rightarrow medical \rightarrow Chest) - a(ImageNet \rightarrow medical \rightarrow natural \rightarrow Chest)$.
These results are presented in Figure~\ref{fig:2_brd_mstl}.

Unlike the two-Bridge TL case there is no apparent pattern towards {\em ImageNet $\rightarrow$ general $\rightarrow$ similar $\rightarrow$ Chest} over {\em ImageNet $\rightarrow$ similar $\rightarrow$ general $\rightarrow$ Chest} in the case of small dataset sizes (5\% or 10\%) with a 13/12 split in each case. This could be due to the dominating factor of ImageNet over the intermediary datasets. Whilst for larger target datasets (15\% and 20\%) {\em ImageNet $\rightarrow$ general $\rightarrow$ similar $\rightarrow$ Chest} starts to dominate the better choices -- with 16 and 19 combinations respectively. Although, as before, the extreams become less.
In addition, the highest accuracies were achieved when using the Covid dataset as the second intermediate dataset. Supporting the case that significantly similar datasets are the best. 

\subsection{Overall benefit of TL}
We bring together the results of the different TL approaches here. Table \ref{tab:best} highlights the best accuracy we were able to achieve at different volumes of the Chest dataset. It is seen that the best accuracy value is obtained with traditional TL at 5\% and 15\% data volume and 3-bridge TL at 10\% and 20\% data volume. It can be said that it is more advantageous to choose traditional TL or 3-bridge TL rather than 2-bridge in order to achieve high performance by using a small volume of data for this case.

\begin{table}
    \centering
        \caption{Best accuracy and TL approach taken}
    \label{tab:best}
    \begin{tabular}{ccc}
        \hline
        Data Volume & Best Accuracy & Approach \\\hline
        5\%   & 91.04\% & Single Stage TL\\
        10\%  & 92.86\% & Three-Bridge TL  \\
        15\%  & 93.17\% & Single Stage TL \\
        20\%  & 93.98\% &  Three-Bridge TL\\
        \hline
    \end{tabular}
\end{table}

\section{Conclusion}
\label{sec:conc}
We set out to determine the validity of two of the three main adages used when talking about improving the results in Deep Learning: more data leads to better results and Transfer Learning leads to better results. We conduct this using twelve datasets trained on a ResNet18 network. Although other networks may yield better results we fix on ResNet18 so that we can compare other changes. 

For the adage more data leads to better results we trained our ResNet18 network from scratch with subsets of the training data from 5\% to 100\% in steps of 5 to determine the relationship between data volume and accuracy. In all cases there was positive correlation between data volume and accuracy -- though there were some times when the accuracy dropped slightly with dataset volume increase. However, there was not a constant rate of increase in accuracy in comparison to volume, but rather a sharp rise in accuracy before reaching $\sim$20-30\% of the total data after which the rate of accuracy improvement became minimal. Thus suggesting that collecting further data after this elbow point may not be the best option to improve accuracy.

For the adage of TL improves performance we try single, double and triple stage TL with the final target being the Chest dataset. For small volumes of the Chest dataset [5\%, 20\%] we did find that Transfer Learning was beneficial. However, for single stage TL we identify that using TL with the wrong source dataset can lead to worse results than just training on the target dataset. One may assume that the closer the source dataset is to the target dataset (as in Covid and Chest) then the more benefit one can get. However, tuberculosis, another X-ray set of the chest was unable to match the accuracy of the Chest dataset when used as a source dataset in Transfer Learning. And could be beaten by radically different datasets such as Animal, Natural and Imagenette.

For 2-bridge Transfer Learning there is no clear winner as to whether you should do a general dataset as the source training dataset followed by a more similar dataset to the target dataset or vice versa. There's a slight advantage in doing general $\rightarrow$ similar $\rightarrow$ target which is strongest when the target set is the smallest, though this is not enough to make a rule. Likewise for 3-bridge TL, there is no clear rule.

Although our results here only demonstrate these patterns for our twelve selected datasets, and final target of Chest, we suspect that they would generalise to other data domains and targets. In the future we seek to expand our results by targeting different datasets, expand to other Deep Learning networks and consider such things as data augmentation. The concept of which datasets are conducive to improve the results for Transfer Learning is an open research question which needs addressing. 

\bibliographystyle{splncs04}
\bibliography{references}

\end{document}